\documentclass{article}

\usepackage{arxiv}

\usepackage[utf8]{inputenc} 
\usepackage[T1]{fontenc}    
\usepackage{hyperref}       
\usepackage{url}            
\usepackage{booktabs}       
\usepackage{amsfonts}       
\usepackage{nicefrac}       
\usepackage{microtype}      
\usepackage{lipsum}		
\usepackage{graphicx}
\usepackage[square,sort,comma,numbers]{natbib}
\usepackage{doi}

\title{Smart Agriculture : A Novel Multilevel Approach for Agricultural Risk Assessment over Unstructured Data}

\date{} 					

\author{
 Hasna Najmi \\
  School of Information Sciences\\
  Meridian Team\\
  LyRICA Laboratory \\
   \And
 Mounia Mikram \\
  School of Information Sciences\\
  Meridian Team\\
  LyRICA Laboratory \\
  \And
 Maryem Rhanoui \\
  School of Information Sciences\\
  Meridian Team\\
  LyRICA Laboratory \\
  \And
 Siham Yousfi \\
  School of Information Sciences\\
  Meridian Team\\
  LyRICA Laboratory \\
}

\renewcommand{\headeright}{}
\renewcommand{\undertitle}{}
\renewcommand{\shorttitle}{Smart Agriculture}

\begin{document}
\maketitle

\begin{abstract}
Detecting opportunities and threats from massive text data is a challenging task for most. Traditionally, companies would rely mainly on structured data to detect and predict risks, losing a huge amount of information that could be extracted from unstructured text data. Fortunately, artificial intelligence came to remedy this issue by innovating in data extraction and processing techniques, allowing us to understand and make use of Natural Language data and turning it into structures that a machine can process and extract insight from. Uncertainty refers to a state of not knowing what will happen in the future. This paper aims to leverage natural language processing and machine learning techniques to model uncertainties and evaluate the risk level in each uncertainty cluster using massive text data.
\end{abstract}


\section{Introduction}

For a sector such a as agriculture, unmanaged risks can potentially cause serious consequences. That goes as far as creating extensive financial and economic losses nationwide and worldwide. In fact, agricultural risks stand at the heart of the biggest crisis the world faces nowadays: food insecurity. The increasing amount of uncertainties that stakeholders in agriculture have to deal with are immense and overwhelming \cite{thornton2014climate, shoaib2021quantifying}.

Agriculture has been known throughout the years to be the backbone of global economy. And thus, in order to increase economic growth, ensure food security and reduce poverty, it is crucial that we innovate strategies to deal with the ever-changing environment in which agriculture operates.
Fortunately, human evolution has paved the way for bigger possibilities due to the industrial technology waves. Many fields, among which agriculture, benefit largely from what technology has to offer in order to improve their end results \cite{bannerjee2018artificial, jha2019comprehensive, talaviya2020implementation}.

The integration of artificial intelligence \cite{patricio2018computer, zhang2021nanotechnology} and data science \cite{zhu2018deep} has revolutionized the practice of agriculture and improved its precision and accuracy.  Known as smart agriculture \cite{gondchawar2016iot, da2021overview}, smart farming \cite{saiz2020smart, ouafiq2021iot} or agriculture 4.0 \cite{rose2018agriculture, zhai2020decision}, this discipline offers many advantages and unprecedented analytical potential, notably in agricultural risk assessment.

Managing risk in agriculture faces two types of problems: anticipated events and unplanned events. Anticipated problems can be assessed and managed systematically according to policies while unplanned events can be challenging as they limit the assessment process and its application.

To identify and monitor uncertainties that matter, the majority of research dealing with risk in the agricultural sector focuses only on quantitative data. The latter is numerical and structured and lack accessibility.

It has been proven that textual and unstructured data embeds non-trivial latent knowledge that can provide significant value for analysis \cite{mooney2005mining, rhanoui2020hybrid, chen2019agrikg}. In fact, projections from International Data Corporation (IDC) predict that, by 2025, almost 80\% of worldwide data will be unstructured. And for numerous large companies, that critical mass has been already reached. 
When structured data fails to provide the analysis needed to identify trends in the agricultural sector, risk managers seek the information in news and text documents. That task requires significant time and effort.

Extracting useful information and combining knowledge from multiple and heterogeneous data sources improves the extracted insight as it enhances completeness and reliability \cite{yousfi2019towards,yousfi2022smart}. However this process is challenging as it involves gathering data from several sources, analysing and understanding unstructured data in natural language in order to identify and classify risk clusters and finally evaluating and assessing the uncertainties.

This paper proposes a solution that leverages unstructured data to identify and model uncertainties in an effective and cost-efficient manner. It focuses on detecting and modeling uncertainties derived from unstructured text data. 

The remainder of this paper is organized as follows: section two presents the principles and basics of risks management, section three introduces some relevant natural language processing techniques, section four discusses the related works, section five presents the proposed approach and finally the last section discusses the results.

\section{Classification and Assessment of Agricultural Risk Management}

According to the Project Management Institute's PMBOK, risk is “an uncertain event or condition that if it occurs has a positive or negative effect on a project’s objectives” \cite{stackpole2013user}. The APM Body of Knowledge defines it as “an uncertain event or set of circumstances that, should it or they occur, would have an effect on achievement of one or more project objectives”. And the Institution of Civil Engineers RAMP6 describes risk as “a possible occurrence which could affect (positively or negatively) the achievement of the objectives of the investment”. All these definitions encompass two concepts that Dr. David Hillson simplified as: “uncertainty that matters”.


Agricultural activity has always been prone to a wide range of uncertainties. Due to the nature of the environment in which it operates, evaluating and managing uncertainties in agriculture is challenging in practice as it is in theory. Conventionally, we distinguish between four key components in Agricultural Risk Management: Risks, Stakeholders, Strategies and Interventions.

The identification, assessment and management of these components is not linear in nature, a holistic and integrated perspective is needed to better illustrate the dynamicity between them \cite{WorldBank2016}.

The following focuses on discovering two of the four components: types of risks and stakeholders; the other two components are associated with finding solutions, which won't be addressed in this study.


In general, the literature in the field of agricultural risks focuses on a limited set of sources in identifying risks, with a slight difference in terms of classification.

OECD Secretariat proposed a presentation of agricultural risks, combining two dimensions \cite{OECD2009} (as shown in table \ref{tab:risks}). This presentation has been adapted from approaches given by various authors \cite{Newbery1989,Harwood1999,Hardaker2004}.

The World Bank, on the other side, classified the types of risk into three categories \cite{WorldBank2016}, as shown in table \ref{tab:wb}.

\begin{table}[!htb]
\footnotesize
\centering
\caption{Classification of types of risks in agriculture according to OECD}
\label{tab:risks}
\begin{tabular}{|p{2cm}|p{3cm}|p{3.5cm}|p{4cm}|}
\toprule
\textbf{Type of risk}         & \textbf{Micro (idiosyncratic) risk affecting an individual household}                 & \textbf{Meso (Covariant) risk affecting groups of households of communities} & \textbf{Macro (Systemic) risks affecting regions or nations}                                        \\ \midrule
\textbf{Market/ Prices}       &                                                                           -         & Changes in price of land, new requirements from food industry                & Changes in input/output prices due to shocks, trade policy, new markets, endogenous variability ... \\ \midrule
\textbf{Production}           & Hail, frost, non-contagious diseases,  personal hazards (illness, death) assets risks & Rainfall, landslides, pollution                                              & Floods, droughts, pests, contagious diseases, technology                                            \\ \midrule
\textbf{Financial}            & Changes in income from other sources (non-farm)                                       &          -                                                                    & Changes in interest rates/value of  financial assets/access to credit                               \\ \midrule
\textbf{Institutional/ Legal} & Liability risk                                                                        & Changes in local policy or regulations                                       & Changes in regional or national policy and regulations, environmental law, agricultural payments    \\ \bottomrule
\end{tabular}
\end{table}

\begin{table}[!htb]
\footnotesize
\centering
\caption{Classification of types of risks according to World Bank}
\label{tab:wb}
\begin{tabular}{|p{2cm}|p{6cm}|p{5cm}|}
\toprule
\textbf{Type of risk}                                                         & \textbf{Examples}                                                                                                                                                             & \textbf{Sections affected}                                                                                                                               \\ \midrule
\textbf{Production Risks}                                                     & Nonextreme weather events, catastrophic weather events, Outbreaks of pests and diseases, damage caused by animals, fire and wind, Human-induced problems (theft, arson, etc), & Yields, supply chain, product quality, disruption in the flow of goods and services.                                                                     \\ \midrule
\textbf{Market risks}                                                         & Prices for inputs and outputs, local and global supply and demand, exchange rate, interest rate, counterparty                                                                 & Price, quality, availability, access to necessary products and services                                                                                  \\ \midrule
\textbf{Enabling environment risks} & Changes in government or business regulations, changes in the macroeconomic environment, political risks, conflict, trade restrictions, logistics, corruption.                & Decision-making, productivity, market options, changes in yields quantity and quality, disruptions in the flow of goods, services, information, finance. \\ \bottomrule
\end{tabular}
\end{table}

To ensure relevance and pertinence, it is important that we keep sources in check. Stakeholders in agriculture have an influence that spreads wide and far. Disclosing information regarding risks and constraints begins with identifying the possessors of that information. Hence, it is crucial to uncover the main actors in the sector of agriculture and link their relationship to the risks mentioned formerly.

The World Bank classifies the stakeholders into three major categories \cite{WorldBank2016}, as shown in table \ref{tab:actors}.

\begin{table}[htbp]
\small
\centering
\caption{Different stakeholders/actors in agriculture}
\label{tab:actors}
\begin{tabular}{|p{2.5cm}|p{5cm}|}
\toprule
\textbf{Stakeholder}           & \textbf{Types}                                                                                     \\ \midrule
Producers                      & Operators of marginal, small, medium-size and large-scale farms                                    \\ \midrule
Commercial sector stakeholders & Commercial stakeholders, traders, wholesalers, retailers, financial institutions, input providers. \\ \midrule
Public sector                  & Public sector institutions, parastatals, government, donors                                        \\ \bottomrule
\end{tabular}
\end{table}

Having identified the fundamentals for the risk study, we proceed to the second stage of risk management: risk assessment. The scientific literature mainly differentiates between two types of models: quantitative and qualitative risk assessment. Qualitative models are techniques which focus on subjective assessment of outcomes: causes and consequences, while quantitative models concentrate on numerical values and the calculation of probabilities.

There are several methods used to measure and evaluate risks, each is suitable for a certain type(s) of risk: What if, Fuzzy matrix \cite{Skorupski2017}, Scenario analysis  \cite{Kosow2008}, Event Tree Analysis (ETA) \cite{Ferdous2011}, Fault Tree Analysis (FTA) \cite{Ferdous2011}, Delfi Technique \cite{Linstone1975}, Monte-Carlo simulation \cite{Rubinstein2016}, Cost-benefit analysis (CBA) \cite{Boardman2017}, Value-at-Risk (VAR) \cite{Jorion2001} and Variation-covariation method.

These methods usually require a considerable amount of historical statistical data and an extensive in-depth knowledge of the nature of each of the risks. Most of them rely on stakeholders and experts for information acquisition, which, despite all efforts to bring objectivity, prove to have a bias in prioritizing certain types of risks over others.

Having covered the basic concepts surrounding risk and risk assessment in agriculture, the next step is to clarify the notions related to the technical side of this study, which are related to Machine Learning algorithms used for unstructured data.

\section{Natural Language Processing}

Unstructured data is data that is unorganized and raw, it doesn’t adhere to a conventional data model or schema and thus cannot be stored in a relational database. It can be both textual or non-textual, and human or machine generated.

Over the past few years, many tools in modern analytics have emerged to solve this struggle and help transform unstructured data into actionable intelligence. One of which is text analytics, using Natural Language Processing (NLP) and Machine Learning (ML).


Machines rely on formal languages to represent information, whereas, we, as humans, rely on multiple criteria to represent and interpret information. NLP is a subfield of computer science and artificial intelligence which studies computational techniques to enable machines to understand and communicate in human language. The applications of NLP can either be used directly or serve as subtasks to larger tasks.

In this study we relied on analyzing our text data following three criteria: syntax (Grammar, Word Order, Word Endings, Speech) semantics (Word meanings, How words link together, Sequencing) and pragmatics (Context, Conversation, Social).

The following sections however will focus on enumerating mainly the algorithms and techniques that we used to perceive information on a semantics and pragmatics levels as the syntax level is pretty standard and similar to previous researches.

\subsection{Part-Of-Speech Tagging}

Part-Of-Speech (POS) tagging \cite{gimpel2010part, owoputi2013improved, plank2016multilingual, mishra2019multi} is the process of tagging a word in a document to its grammatical category that it serves within the sentence.

\subsection{Question Answering}

“Question Answering is a fundamental problem in Natural Language Processing where we want the computer system to be able to understand a corpus of human knowledge (context) and answer questions provided in Natural Language” \cite{Jain2017}.

The goal behind a QA model is to compute a system that would be able to understand a passage and answer a question about it \cite{ravichandran2002learning, antol2015vqa, yang2016stacked, kwiatkowski2019natural}.

The most recent advances in neural networks and deep learning have paved the way for QA models to evolve achieving significant results in terms of performance metrics. Most of these models rely on Stanford Question Answering Dataset (SQuAD) for training.

The SQuAD was published by Rajpurkar et al. \cite{Rajpurkar2016}. The data set contains +100K training examples. Each example is a triplet of three elements:
\begin{itemize}
\item Context: a document or paragraph in natural language (English)
\item Question: a question about the context written in English
\item Answer: a ground-truth (label or target) answer about the aforementioned question.
\end{itemize}

The specificity of SQuAD is that all the answers are supposed to be continuous spans of the input context. It means that no matter the question, the answer will be a group of consecutive words from the input paragraph. This means that in order to apply a model trained on SQuAD, we have to be sure the answer to the question we are asking actually lies within the context paragraph and it is continuous and not fragmented.

To sum up, a model trained on SQuAD takes as input a paragraph and a question and outputs the most likely span of the paragraph that could serve as answer to the given question.

Reading comprehension problem formulation:
\begin{itemize}
\item \textbf{Input}: a training set $D$ of $N$ examples

$D=(p_i,q_i,a_i)_{i\in\{1,\ldots,N\} }$

where $p_i$, $q_i$ and $a_i$ respectively denote the context paragraph, the question and the target answer of the $ith$ example.
\item \textbf{Output}: a function $f$ that takes as input a context paragraph $p$ and a question $q$ and returns an answer $a = f_\Theta(p, q)$ from paragraph $p$ that maximizes the expected log-probability $\mathbb{E}_{(p,q,a)\sim D}[log(\mathbb{P}(a|p,q,\Theta))]$
\end{itemize}

\subsection{Latent Dirichlet Allocation}

Topic modelling algorithms are used to extract a set of topics from a large collection of documents. Latent Dirichlet Allocation (LDA) is an unsupervised machine learning technique and one of the common probabilistic topic models using posterior inference, to analyze large collections of documents with the aim to discover the latent topics that pervade the texts \cite{beli2003latent}.

The intuition behind LDA model is that each document exhibits multiple topics, through a set of rules. The reality is however different as the only observable features we have are the documents and the words appearing in each of them. The other parameters are hidden. The goal of LDA is to infer the latent variables by computing their distribution conditioned on the documents, based on a set of given rules which are called “hyper-parameters”.

\subsection{Sentiment analysis}

Text information can be categorized into two main types: facts and opinions. Facts consist of objective reality, while opinions are subjective views and judgements that describe sentiments and feelings formed toward a certain topic. Opinions can either be direct or comparative, and explicitly expressed or implicitly implied in a neutral sentence.

Sentiment analysis or Opinion Mining \cite{Liu2012} is the automated process of identifying and extracting opinions about a given subject from text data. It can be applied on complete documents (full paragraphs), single sentences or sub-expressions within a phrase \cite{lin2009joint, dos2014deep, bakshi2016opinion}.

Sentiment analysis can be modeled as a classification problem with two sub-tasks:

\begin{itemize}
\item Subjectivity classification: where the model determines whether a sentence is subjective or objective.

\item Polarity classification: where the model determines whether the opinion expressed in the sentence is positive, negative or neutral.
\end{itemize}

For sentiment analysis for this project, we choose to work with polarity classification, and more specifically: Valence-Aware Dictionary for sentiment Reasoning (VADER). We present the algorithm below. 

\subsubsection{VADER Sentiment Analysis}

Valence-Aware Dictionary for sentiment Reasoning is a rule-based “gold-standard sentiment lexicon that is especially attuned to microblog-like contexts. It combines quantitative and qualitative methods that embody grammatical and syntactical conventions for expressing and emphasizing sentiment intensity” \cite{Hutto2014}.

VADER was originally accommodated to work on micro-blog like contexts, yet it smoothly generalizes to multiple domains and outperforms most of the traditional Sentiment Analysis methods.

Based on a humanly crafted valence-based sentiment lexicon, VADER doesn’t require any training data, and thus allows a speed-up gain in terms of performance especially when applied to online streaming data.

\subsubsection{VADER scoring method}

Given a certain text, VADER returns the following:

\begin{itemize}
\item Compound score (CS): it’s a uni-dimensional measure of a “normalized, weighted composite score”. It is computed by calculating the sum of all the valence-based scores of each word in the lexicon, and then normalized between -1 (most extreme negative) and +1 (most extreme positive).
\item Positive, negative and neutral scores: they are multidimensional measures of sentiment that represent the ratios for proportions of the text that fall in each of those categories. The positive, negative and neutral scores add up to be 1.
\end{itemize}

This scoring method incorporates quantification derived from the impact of grammatical and syntactical rules described in \cite{Hutto2014} to compute the intensity of sentiment on a sentence-level text.

\section{Related Works}

Risk Management is majorly reliant on the usage of quantitative data. However, over the past few years, researchers have started to shift their attention toward unstructured data to extract analytical insights for intelligent data-driven decision making.

The inclination toward the use of textual data is mostly present within the financial and economical fields \cite{ozbayoglu2020deep, necba2018using, colladon2020forecasting}. Researchers have started to depend on text mining techniques and semantic approaches to model additional risks such as determining the predictability of financial markets through the interpretation of sentiment in online social media and news \cite{Nassirtoussi2015}, and forecasting risk and return in stock markets by incorporating Natural Language Processing algorithms to classify news articles content \cite{Calomiris2019}.

The most recent works attempting to leverage unstructured data have undertaken a different tactic in risk modeling by combining different NLP models at once to improve performance, mainly through constructing features.

Tai \cite{Tai2018} introduced a detailed multi-factor sentiment analysis method to evaluate the volatility of the financial market from intangible data, using NLP algorithms such as sentiment analysis (Loughran-McDonald, Harvard GI), similarity (Jaccard, Cosine), readability (Fog, Flesch) and topic modeling (LDA).

On the other hand, Wahyudin \cite{Wahyudin2016} based his risk analysis methodology on clustering documents based on term presence using K-Means algorithm. He then, in order to quantify the risk level within each cluster, merged the sentiment score (SentiWordNet 3.0) with the term importance (TF-IDF). Finally, to evaluate the clusters quality, he used Silhouette function and Sum of squared error. As for sentiment analysis evaluation, he performed manual analysis.

Tai \cite{Tai2018} relied mainly on sentiment analysis to study volatility and the changes within the texts, while Wahyudin \cite{Wahyudin2016} relied on document clustering to group together documents that discuss the same type of risk.

Colladon et al. \cite{colladon2020forecasting} propose a new textual data index extracted from Italian press capable of predicting the Italian stock market in the COVID-19 crisis and demonstrate the improvement using a forecasting model.

Although the state-of-the-art solutions focus mainly on risk analysis in the financial and economical fields, we can still convey that knowledge and adapt it to suit the agricultural risk management system. In fact, the very nature of the field is of little importance when dealing with solutions that aim to extract insight from unstructured data. Machine learning algorithms operate differently from the human mind as they rely on statistical methods to understand text data, which allows them to maintain uniformity even when the context changes.

What previous works failed to bring is a way to evaluate the validity of the sentiment scores of topics that is somehow automated. This research brings that layer of validation by making use of Question Answering, which allows for the analyst to make sure his risk detection or score is reviewed.

\section{Proposed Approach for Uncertainty Extraction, Scoring and Evaluation}

The objective of the proposed model is to extract all topics from a database containing news articles and papers regarding African Agriculture, these topics will be called “uncertainties”. Next, in order to assess which of these uncertainties matter and to which degree, Sentiment Analysis will be used to evaluate the sentiment in each of the topics. Finally, to be able to interpret the results and analyze them, the model relies on QA algorithm to check the exact answers to the formulated questions regarding each uncertainty.

As opposed to the financial field, the agriculture field doesn’t have enough open source data to allow deep extraction and analysis of uncertainties.

\begin{figure}[!htbp]
  \centering
\centerline{\includegraphics[width=5in]{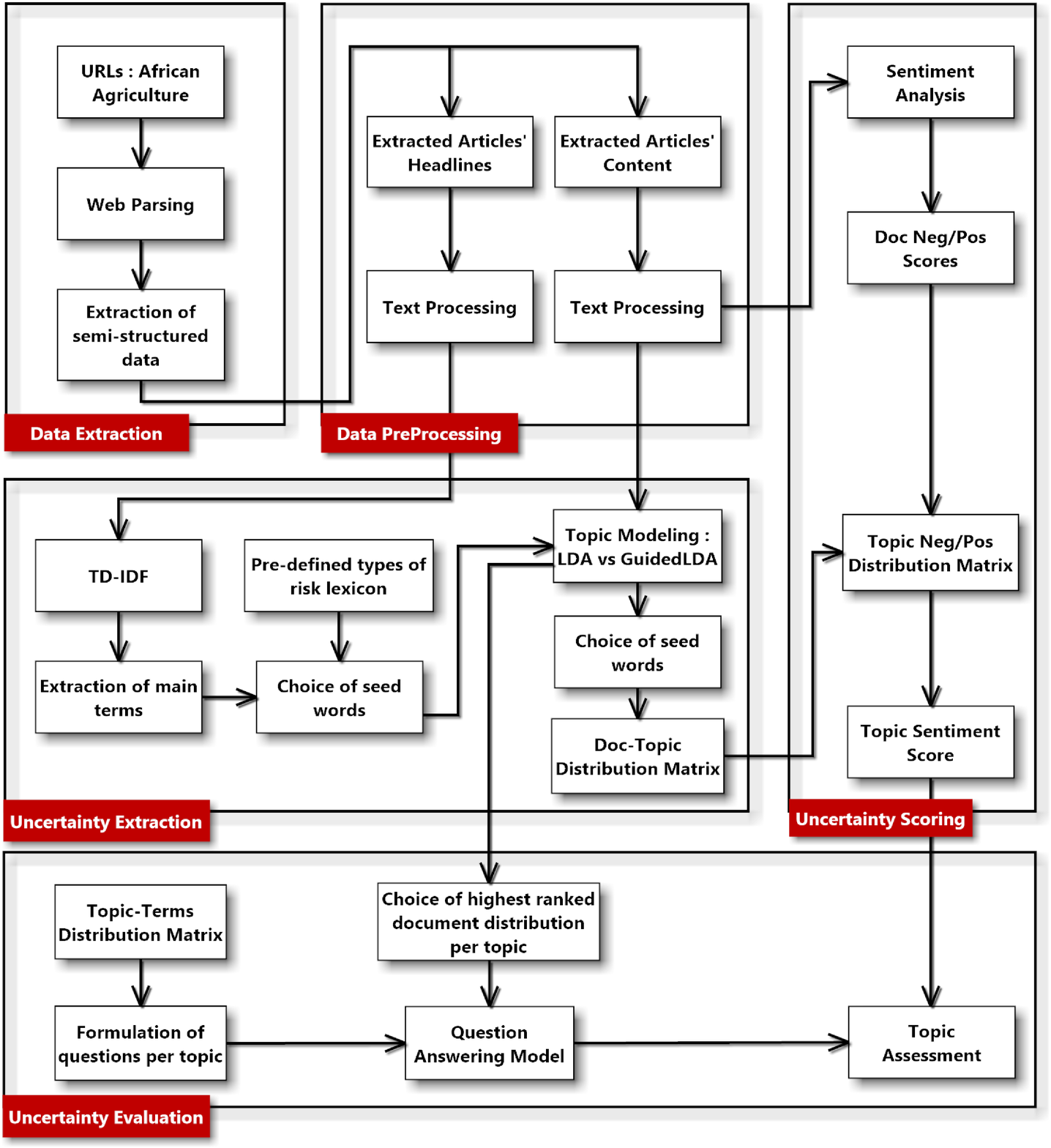}}
  \caption{Proposed Model}
\label{fig:model}
 \end{figure}

In this paper, the structure of the proposed solution \ref{fig:model} includes five main phases; Data Extraction, data pre-processing, uncertainty extraction, uncertainty scoring and uncertainty evaluation.

The rest of the article will go into more detail on each of these steps of the whole process.

\subsection{Data Extraction and Pre-processing}

This study depended mainly on information available online and in English.

The final choices of sources were:

\begin{itemize}
\item Scidev.net: An independent news network operated by Centre for Agriculture and Biosciences International (CAB) which is a not-for-profit organization established under a United Nations treaty agreement. It’s a source of reliable and authoritative news, views and analysis on information about science and technology for global development.
\item The Conversation Africa: An independent source of news and views from the academic and research community. It consists of a team of professional editors working with university and research experts to unlock their knowledge for use by the public.
\end{itemize}

This paper relies on extracting the title and content of articles from 2015 to 2019. 

For the pre-processing of data, this study started off with cleaning the data through processes during which incomplete, incorrect, inaccurate or irrelevant parts of the data are replaced, modified or deleted. Second, data is combined from both sources into a uniform representation under one csv file. The step that comes after that is the data transformation which is converting the data from one format to another through the application of multiple processes: case normalization, tokenization and segmentation, lemmatization and noise removal (such as punctuation and stop words).

\subsection{Uncertainty extraction}

This phase is the focal point of the proposed solution.For feature extraction, we use Bag of Words (BoW) model and for topic modeling, we rely on LDA . 
BoW is a representation of text that describes the occurrence of words within a document. 
Next, three approaches are ran :LDA, LDA wih TF-IDF, and Guided LDA, on the pre-processed content using BoW to explore the terms occurring in each extracted topic and their relative weights. To keep only domain specific content, we run TF-IDF to weigh down the frequent terms while scaling up the rare ones.

\begin{figure}[htbp]
  \centering
\centerline{\includegraphics[width=0.8\textwidth]{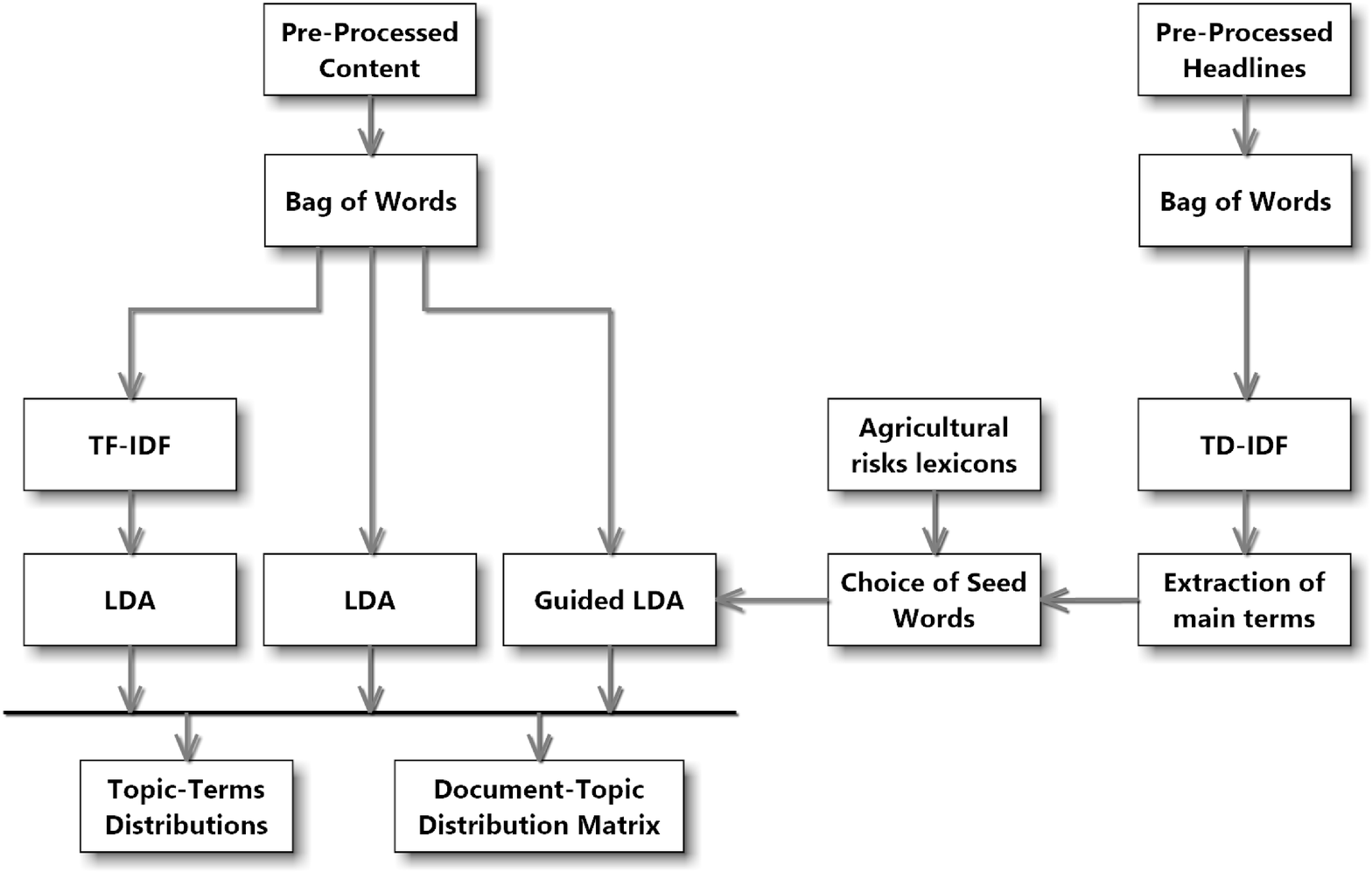}}

  \caption{The process of uncertainty extraction}
\label{fig:uncer}
 \end{figure}

The study then tries a different approach where it doesn't just explore the documents, but also guides the exploration towards the distinctions that we find to be more interesting. Namely, we want to provide the model with seed words that are representative of the articles. A way to do so is to use the information in the headlines. As shown in Figure \ref{fig:uncer}, we will extract the main terms from the headlines using TF-IDF. And use the agricultural risk tables from section 2 to extract a lexicon. These will be the discriminating features that we feed to the LDA model as initial sets of seed words.

The final output of this phase is a topic-term distribution matrix and a document-topic distribution matrix. The probability distribution of the topics in the article is accounted for by the Document-Topic matrix. Similarly, the Topic-Terms matrix accounts for the probability distribution of words derived from that topic.

\subsection{Uncertainty scoring}

At this stage, having distinguished the uncertainties that dominate each document and their respective distributions, what's needed is to extract the sentiment of each of the extracted topics. A negative or positive sentiment would mean that the uncertainty matters and it should be considered as a risk, while a neutral sentiment can have us discard the uncertainty extracted.

First, we go back to our unprocessed content file, and tokenize each document by sentence.

\begin{figure}[htbp]
  \centering
\centerline{\includegraphics[width=0.8\textwidth]{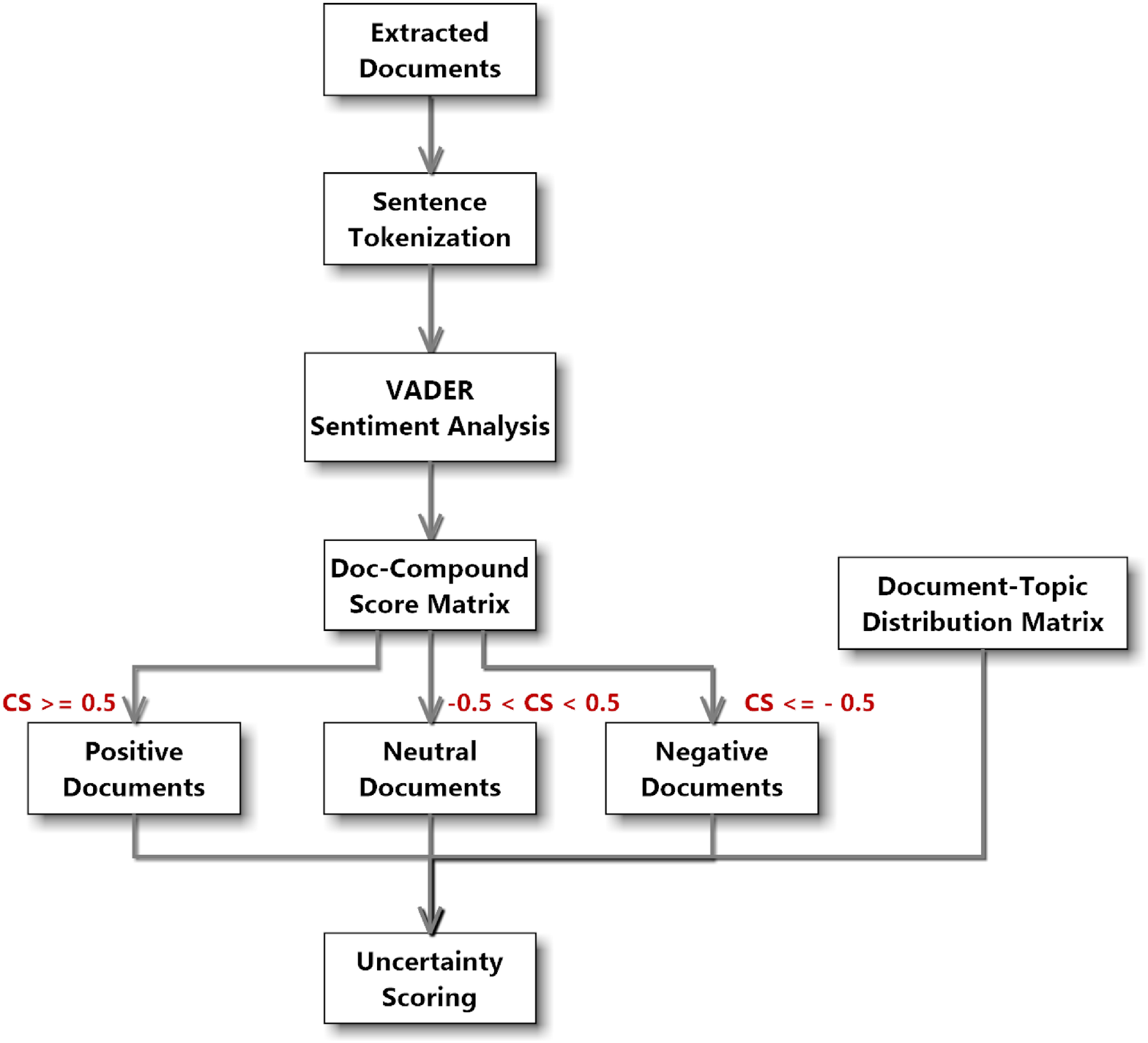}}
  \caption{The process of Uncertainty Scoring}
\label{fig:cs}
 \end{figure}

Second, we run VADER Sentiment Analyzer on each of the documents. The model shall return a positive, negative, neutral and CS per document. We cluster the documents based on their CS as shown in the Figure \ref{fig:cs}.

Afterward, we’ll go back to the LDA output: Document-Topic Distribution Matrix, we’ll keep the distribution of the topic that most represents a document and discard the other distributions. Then, we’ll cluster together the documents that share the same topic, and compute the sentiment score (SS) of each topic.

\subsection{Uncertainty evaluation}

The uncertainty evaluation is the last step of the proposed solution. To ensure that proper sentiment scores were returned, QA model will be used.

This phase requires human intervention at two levels: first in formulating the questions and choosing the documents that we input to the QA model, and second in the interpretation of the answer obtained by the QA model.

The study will rely on topic-terms distribution matrix from the LDA model to formulate the questions. The answer provided by the QA model will allow the interpretation of the sentiment score and the evaluation of the results given in the previous phase.

Finally, the final results will be organized combining expert intervention with the models’ results to produce a proper analysis on the uncertainties extracted from the data.

\begin{figure}[!h]
  \centering
\centerline{\includegraphics[width=5in]{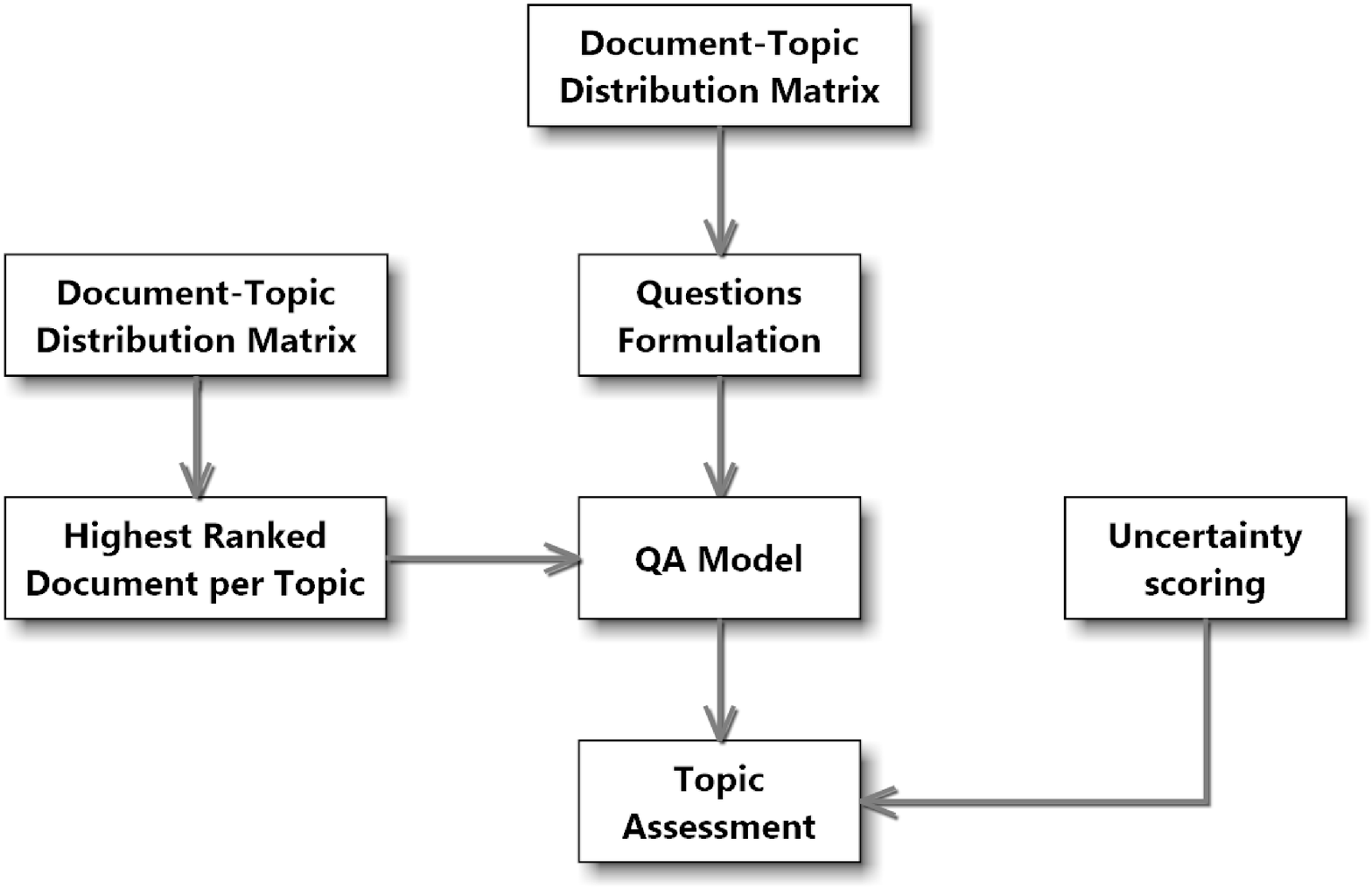}}
  \caption{The process of uncertainty assessment phase}
\label{fig:uev}
 \end{figure}

\section{Results and Discussion}

\subsection{Topic Modeling}

For the topic modeling part, we started with the conventional LDA model.
First, we use BoW model. We create a dictionary from “text-list” containing the number of times a word appears in the training set. Then we filter out rare words that appear in less than 15 documents and common words that appear in more than 90 percent of the documents. Finally, we kept the first 3000 most frequent terms. We then run LDA model on the BoW output. We set the number of topics to 6 and the number of words to appear per topic to 10.

Second, we apply TF-IDF transformation to the entire corpus. Afterward, we run LDA model. We run and compare multiple variants of LDA models (LDA, LDA with TF-IDF and GuidedLDA).LDA with TF-IDF showed the best performances the results  are shown in Table \ref{tab:tfidf}.

\begin{table*}[!h]
\footnotesize
\centering
\caption{LDA applied on TF-IDF (Topic-Term Distributions)}
\label{tab:tfidf}   
\begin{tabular}{@{}lllllll@{}}
\toprule
\textbf{LDA-TF-IDF} & \textbf{Topic 0} & \textbf{Topic 1} & \textbf{Topic 2} & \textbf{Topic 3} & \textbf{Topic 4} & \textbf{Topic 5} \\ \midrule
\textbf{Word 1}  & seed             & data             & climate          & irrigation       & land             & cassava          \\
\textbf{Word 2}  & disease          & fishery          & change           & innovation       & reform           & loss             \\
\textbf{Word 3}  & city             & seed             & tree             & land             & water            & fall             \\
\textbf{Word 4}  & variety          & soil             & specie           & soil             & climate          & disease          \\
\textbf{Word 5}  & water            & land             & rainfall         & insect           & farm             & study            \\
\textbf{Word 6}  & technology       & smallholder      & group            & youth            & household        & weather          \\
\textbf{Word 7}  & maize            & conservation     & plant            & woman            & energy           & research         \\
\textbf{Word 8}  & wheat            & water            & policy           & report           & system           & scientist        \\
\textbf{Word 9}  & pest             & information      & growth           & initiative       & trade            & potato           \\
\textbf{Word 10} & woman            & farmer           & seed             & reform           & change           & variety          \\ \bottomrule
\end{tabular}
\end{table*}

In this particular case, LDA using TF-IDF returned interpretable results compared to other methods. 
From a sector specific point of view, we labeled the extracted topics as shown in Table \ref{tab:topicnam}.
\begin{table}[!h]
\small
\centering
\caption{Labeled topics list}
\label{tab:topicnam}
\begin{tabular}{@{}lllllll@{}}
\toprule
\textbf{Topic number} & \textbf{Topic name}\\ \midrule
\textbf{Topic 0}     & {Agricultural technical innovation}  \\
\textbf{Topic 1}     & {Weather conditions}  \\
\textbf{Topic 2}     & {Economical ecosystem}  \\ 
\textbf{Topic 3}     & {Land management}  \\
\textbf{Topic 4}     & {Geological optimization}  \\
\textbf{Topic 5}     & {Crop mixes}  \\
\bottomrule
\end{tabular}
\end{table}
We carried on the next steps using the outcome from the LDA using TF-IDF model. 

Thus, we extracted the Document-Topic Distributions of the LDA using TF-IDF model. The results of this computation are shown in Table \ref{tab:dist}. Each line gives us the distribution of each of the topics in the columns per document. And so for each pair (a, b): "a" refers to the topic's number on that column, and "b" refers to the distribution of that same topic in the document ID of that same line. 

\begin{table}[!h]
\small
\centering
\caption{A snippet from the Document-Topic Distribution results running LDA on TF-IDF}
\label{tab:dist}
\begin{tabular}{@{}lllllll@{}}
\toprule
\textbf{Doc Id} & \textbf{Topic 0} & \textbf{Topic 1} & \textbf{Topic 2} & \textbf{Topic 3} & \textbf{Topic 4} & \textbf{Topic 5} \\ \midrule
\textbf{0}      & (0, 0.023) & (1, 0.879)   & (2, 0.023) & (3, 0.023) & (4, 0.023) & (5, 0.024) \\
\textbf{1}      & (0, 0.028) & (1, 0.028)  & (2, 0.028) & (3, 0.028) & (4, 0.028)  & (5, 0.858)   \\
\textbf{2}      & (0, 0.027) & (1, 0.862)  & (2, 0.027) & (3, 0.027) & (4, 0.027) & (5, 0.027) \\
\textbf{3}      & (0, 0.860)  & (1, 0.027)  & (2, 0.027) & (3, 0.027) & (4, 0.027) & (5, 0.028)  \\
\textbf{4}      & (0, 0.523)   & (1, 0.019) & (2, 0.020) & (3, 0.019) & (4, 0.019) & (5, 0.396)   \\
\textbf{5}      & (0, 0.019) & (1, 0.019) & (2, 0.019) & (3, 0.019)  & (4, 0.591)  & (5, 0.329)  \\
\textbf{6}      & (0, 0.516)   & (1, 0.020) & (2, 0.020) & (3, 0.020) & (4, 0.020) & (5, 0.402)   \\
\textbf{7}      & (0, 0.018) & (1, 0.636)    & (2, 0.018) & (3, 0.018)  & (4, 0.018) & (5, 0.288)  \\
\textbf{8}      & (0, 0.575)   & (1, 0.022) & (2, 0.022) & (3, 0.022) & (4, 0.022)  & (5, 0.333)   \\
\textbf{9}      & (0, 0.029) & (1, 0.029) & (2, 0.029) & (3, 0.029)  & (4, 0.029) & (5, 0.853)   \\
\textbf{10}     & (0, 0.032)  & (1, 0.031) & (2, 0.031)  & (3, 0.067)  & (4, 0.031) & (5, 0.804)   \\ \bottomrule
\end{tabular}
\end{table}

\subsection{Uncertainty scoring}

For uncertainty scoring, we apply VADER Sentiment analysis on the content file. Then we  extract the compound score relative to each document. Next, we compute the Sentiment Score (SS) for each topic. 

And applying the formula mentioned in the previous section, we obtain the scores for each topic (Table \ref{tab:unsc})
 
\begin{table}[htbp]
\small
\centering
\caption{Uncertainty Score}
\label{tab:unsc}
\begin{tabular}{@{}lllllll@{}}
\toprule
                                                                   & \textbf{Topic 0} & \textbf{Topic 1} & \textbf{Topic 2} & \textbf{Topic 3} & \textbf{Topic 4} & \textbf{Topic 5} \\ \midrule
\textbf{\begin{tabular}[c]{@{}l@{}}SS \\ \end{tabular}} & 0.394084         & 0.0038123        & 0.204868         & 0.031465         & 0.024643         & 0.022602         \\ \bottomrule
\end{tabular}
\end{table}
 
As shown in table \ref{tab:unsc}, Topic 0: Agricultural tech innovation, and Topic 2: Economical ecosystem, have considerable high scores in the positive range which means that they represent opportunities for agricultural domain. As opposed to the other remaining topics (Topic 1, Topic 3, Topic 4, Topic 5) whose scores are closer to the neutral range. Consequently the latest need further context analysis to determine whether they relate to opportunities or risks.
Sentiment Scores are a reflection of word scores used in the document, and so a neutral score given to a certain topic would mean that the words used around that topic have a neutral tone, a positive score given to a certain topic would mean that the words used around that topic have a positive tone, and a negative score given to a certain topic would mean that the words used around that topic are more of negative tone.
Checking the validity of these scores is about going back to the text itself and reading the document, or paragraph that refers to that certain topic. 
This study introduces a method to do so and navigate into the paragraphs in an easier way using the Question Answering approach, which is explained in the following section. 

\subsubsection{Uncertainty evaluation}

To further confirm the uncertainty assessment, the study uses a basic Question Answering approach that, given a paragraph and a question, returns the answer that best responds to the query. This step is optional since it adds a functional evaluation to the results. The sentiment score recalled is sufficient to evaluate the uncertainty.

\begin{figure}[htbp]
  \centering
\centerline{\includegraphics[width=0.5\textwidth]{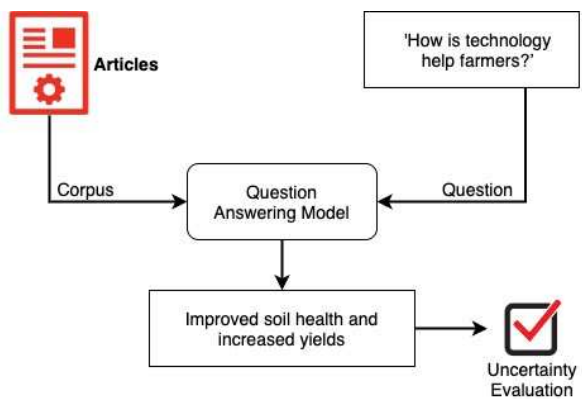}}
  \caption{Uncertainty's evaluation with QA}
\label{fig:eval}
 \end{figure}

We evaluate the high score of Uncertainty of Topic 0: Agricultural tech innovation and confirm the results: 

\begin{itemize}
\item We picked one of the documents that had the highest Document-Topic0 distribution, we input it as a paragraph in the deeppavlov.ai model.
\item We use the Topic-Word matrix and formulate a question which combines words that most represent Topic 0. And finally, we input our question into the model.
\item We compute the answer and observe the results as shown in the figure \ref{fig:eval}
\end{itemize}

What we notice from the figure \ref{fig:eval} is that the answer returned implies a positive sentiment regarding the topic discussed by the document, which aligns with the SS (0,39) that we computed in the previous section.
Summing up, this approach has allowed us to extract uncertainties regarding the agricultural sector from scraped unstructured text data and determine what matters from it through a conduction of sentiment analysis on pulled topics, and proving the correctness of the results through Question Answering models. This type of retrieved information would allow for a dynamic focus on risks by constantly following the trends in the sector as it changes and moves. 

\subsection{Discussion}

As opposed to related works, this project focused more on the uniformity of information extraction instead of information clustering. The agricultural field lacks the suitable lexicon to conduct studies using supervised learning models, and so naturally, what we aimed to achieve is a work where we let the corpus lead the results instead of having a human do it. What the other works also fail to bring to the table is a method of score evaluation of topic scoring. This project introduces Question Answering as an added step to allow the validation of the final Sentiment Scores through analyst intervention.

\section{Conclusion}

This paper proposes a novel approach to extract insight from unstructured data regarding agricultural risks. 

It relied mainly on state-of-the-art Natural Language Processing techniques to extract insight and evaluate it. 
The approach chosen for this study has allowed us to prove the harmonious movement between phases in the model suggested, while maintaining uniformity in analysis.

This model minimizes the human intervention through its various phases by allowing a first-hand analysis run through the algorithms. However, and to check the validity of the conclusions drawn from scores given by the model, each step allows a certain intervention that check the effectiveness of the results.

For future works, we aim to experiment furthermore with data pre-processing techniques in order to yield better results. Another challenge that could surface is in finding efficient and effective methods of adding new information when new risks emerge.

\bibliographystyle{unsrt}  

\begin{thebibliography}{10}

\bibitem{thornton2014climate}
Philip~K Thornton, Polly~J Ericksen, Mario Herrero, and Andrew~J Challinor.
\newblock Climate variability and vulnerability to climate change: a review.
\newblock {\em Global change biology}, 20(11):3313--3328, 2014.

\bibitem{shoaib2021quantifying}
Syed~Abu Shoaib, Mohammad Zaved~Kaiser Khan, Nahid Sultana, and Taufique~H
  Mahmood.
\newblock Quantifying uncertainty in food security modeling.
\newblock {\em Agriculture}, 11(1):33, 2021.

\bibitem{bannerjee2018artificial}
Gouravmoy Bannerjee, Uditendu Sarkar, Swarup Das, and Indrajit Ghosh.
\newblock Artificial intelligence in agriculture: A literature survey.
\newblock {\em International Journal of Scientific Research in Computer Science
  Applications and Management Studies}, 7(3):1--6, 2018.

\bibitem{jha2019comprehensive}
Kirtan Jha, Aalap Doshi, Poojan Patel, and Manan Shah.
\newblock A comprehensive review on automation in agriculture using artificial
  intelligence.
\newblock {\em Artificial Intelligence in Agriculture}, 2:1--12, 2019.

\bibitem{talaviya2020implementation}
Tanha Talaviya, Dhara Shah, Nivedita Patel, Hiteshri Yagnik, and Manan Shah.
\newblock Implementation of artificial intelligence in agriculture for
  optimisation of irrigation and application of pesticides and herbicides.
\newblock {\em Artificial Intelligence in Agriculture}, 2020.

\bibitem{patricio2018computer}
Diego~In{\'a}cio Patr{\'\i}cio and Rafael Rieder.
\newblock Computer vision and artificial intelligence in precision agriculture
  for grain crops: A systematic review.
\newblock {\em Computers and electronics in agriculture}, 153:69--81, 2018.

\bibitem{zhang2021nanotechnology}
Peng Zhang, Zhiling Guo, Sami Ullah, Georgia Melagraki, Antreas Afantitis, and
  Iseult Lynch.
\newblock Nanotechnology and artificial intelligence to enable sustainable and
  precision agriculture.
\newblock {\em Nature Plants}, 7(7):864--876, 2021.

\bibitem{zhu2018deep}
Nanyang Zhu, Xu~Liu, Ziqian Liu, Kai Hu, Yingkuan Wang, Jinglu Tan, Min Huang,
  Qibing Zhu, Xunsheng Ji, Yongnian Jiang, et~al.
\newblock Deep learning for smart agriculture: Concepts, tools, applications,
  and opportunities.
\newblock {\em International Journal of Agricultural and Biological
  Engineering}, 11(4):32--44, 2018.

\bibitem{gondchawar2016iot}
Nikesh Gondchawar, RS~Kawitkar, et~al.
\newblock Iot based smart agriculture.
\newblock {\em International Journal of advanced research in Computer and
  Communication Engineering}, 5(6):838--842, 2016.

\bibitem{da2021overview}
Franco da~Silveira, Fernando~Henrique Lermen, and Fernando~Gon{\c{c}}alves
  Amaral.
\newblock An overview of agriculture 4.0 development: Systematic review of
  descriptions, technologies, barriers, advantages, and disadvantages.
\newblock {\em Computers and Electronics in Agriculture}, 189:106405, 2021.

\bibitem{saiz2020smart}
Ver{\'o}nica Saiz-Rubio and Francisco Rovira-M{\'a}s.
\newblock From smart farming towards agriculture 5.0: A review on crop data
  management.
\newblock {\em Agronomy}, 10(2):207, 2020.

\bibitem{ouafiq2021iot}
El~Mehdi Ouafiq, Abdessamad Elrharras, A~Mehdary, Abdellah Chehri, Rachid
  Saadane, and Mohamed Wahbi.
\newblock Iot in smart farming analytics, big data based architecture.
\newblock In {\em Human Centred Intelligent Systems}, pages 269--279. Springer,
  2021.

\bibitem{rose2018agriculture}
David~Christian Rose and Jason Chilvers.
\newblock Agriculture 4.0: Broadening responsible innovation in an era of smart
  farming.
\newblock {\em Frontiers in Sustainable Food Systems}, 2:87, 2018.

\bibitem{zhai2020decision}
Zhaoyu Zhai, Jos{\'e}~Fern{\'a}n Mart{\'\i}nez, Victoria Beltran, and
  N{\'e}stor~Lucas Mart{\'\i}nez.
\newblock Decision support systems for agriculture 4.0: Survey and challenges.
\newblock {\em Computers and Electronics in Agriculture}, 170:105256, 2020.

\bibitem{mooney2005mining}
Raymond~J Mooney and Razvan Bunescu.
\newblock Mining knowledge from text using information extraction.
\newblock {\em ACM SIGKDD explorations newsletter}, 7(1):3--10, 2005.

\bibitem{rhanoui2020hybrid}
Maryem Rhanoui, Mounia Mikram, Siham Yousfi, Ayoub Kasmi, and Naoufel Zoubeidi.
\newblock A hybrid recommender system for patron driven library acquisition and
  weeding.
\newblock {\em Journal of King Saud University-Computer and Information
  Sciences}, 2020.

\bibitem{chen2019agrikg}
Yuanzhe Chen, Jun Kuang, Dawei Cheng, Jianbin Zheng, Ming Gao, and Aoying Zhou.
\newblock Agrikg: an agricultural knowledge graph and its applications.
\newblock In {\em International Conference on Database Systems for Advanced
  Applications}, pages 533--537. Springer, 2019.

\bibitem{yousfi2019towards}
Siham Yousfi, Maryem Rhanoui, and Dalila Chiadmi.
\newblock Towards a generic multimodal architecture for batch and streaming big
  data integration.
\newblock {\em Journal of Computer Science}, 15(1):207--220, 2019.

\bibitem{yousfi2022smart}
Siham Yousfi, Dalila Chiadmi, and Maryem Rhanoui.
\newblock Smart big data framework for insight discovery.
\newblock {\em Journal of King Saud University-Computer and Information
  Sciences}, 2022.

\bibitem{stackpole2013user}
Cynthia~Snyder Stackpole.
\newblock {\em A User's Manual to the PMBOK Guide}.
\newblock John Wiley \& Sons, 2013.

\bibitem{WorldBank2016}
World Bank.
\newblock Agricultural sector risk assessment : Methodological guidance for
  practitioners.
\newblock Technical report, Washington, DC, 2016.

\bibitem{OECD2009}
Organisation for Economic Co-operation and Development.
\newblock {\em Managing risk in agriculture: a holistic approach}.
\newblock OECD publishing, 2009.

\bibitem{Newbery1989}
David~M Newbery.
\newblock The theory of food price stabilisation.
\newblock {\em The Economic Journal}, 99(398):1065--1082, 1989.

\bibitem{Harwood1999}
Joy~L Harwood, Richard~G Heifner, Keith~H Coble, Janet~E Perry, and Agapi
  Somwaru.
\newblock Managing risk in farming: concepts, research, and analysis.
\newblock Technical report, 1999.

\bibitem{Hardaker2004}
J~Brian Hardaker.
\newblock {\em Coping with risk in agriculture}.
\newblock Cabi, 2004.

\bibitem{Skorupski2017}
J~Skorupski.
\newblock Fuzzy risk matrix as a tool for the analysis of the air traffic
  safety.
\newblock In {\em In 26th Conference on European Safety and Reliability
  (ESREL)}, pages 2781--2788, 2017.

\bibitem{Kosow2008}
Hannah Kosow and Robert Ga{\ss}ner.
\newblock {\em Methods of future and scenario analysis: overview, assessment,
  and selection criteria}, volume~39.
\newblock DEU, 2008.

\bibitem{Ferdous2011}
Refaul Ferdous, Faisal Khan, Rehan Sadiq, Paul Amyotte, and Brian Veitch.
\newblock Fault and event tree analyses for process systems risk analysis:
  uncertainty handling formulations.
\newblock {\em Risk Analysis: An International Journal}, 31(1):86--107, 2011.

\bibitem{Linstone1975}
Harold~A Linstone, Murray Turoff, et~al.
\newblock {\em The delphi method}.
\newblock Addison-Wesley Reading, MA, 1975.

\bibitem{Rubinstein2016}
Reuven~Y Rubinstein and Dirk~P Kroese.
\newblock {\em Simulation and the Monte Carlo method}, volume~10.
\newblock John Wiley \& Sons, 2016.

\bibitem{Boardman2017}
Anthony~E Boardman, David~H Greenberg, Aidan~R Vining, and David~L Weimer.
\newblock {\em Cost-benefit analysis: concepts and practice}.
\newblock Cambridge University Press, 2017.

\bibitem{Jorion2001}
Philippe Jorion.
\newblock {\em Value at risk: the new benchmark for managing financial risk},
  volume~2.
\newblock McGraw-Hill New York, 2001.

\bibitem{gimpel2010part}
Kevin Gimpel, Nathan Schneider, Brendan O'Connor, Dipanjan Das, Daniel Mills,
  Jacob Eisenstein, Michael Heilman, Dani Yogatama, Jeffrey Flanigan, and
  Noah~A Smith.
\newblock Part-of-speech tagging for twitter: Annotation, features, and
  experiments.
\newblock Technical report, Carnegie-Mellon Univ Pittsburgh Pa School of
  Computer Science, 2010.

\bibitem{owoputi2013improved}
Olutobi Owoputi, Brendan O’Connor, Chris Dyer, Kevin Gimpel, Nathan
  Schneider, and Noah~A Smith.
\newblock Improved part-of-speech tagging for online conversational text with
  word clusters.
\newblock In {\em Proceedings of the 2013 conference of the North American
  chapter of the association for computational linguistics: human language
  technologies}, pages 380--390, 2013.

\bibitem{plank2016multilingual}
Barbara Plank, Anders S{\o}gaard, and Yoav Goldberg.
\newblock Multilingual part-of-speech tagging with bidirectional long
  short-term memory models and auxiliary loss.
\newblock {\em arXiv preprint arXiv:1604.05529}, 2016.

\bibitem{mishra2019multi}
Shubhanshu Mishra.
\newblock Multi-dataset-multi-task neural sequence tagging for information
  extraction from tweets.
\newblock In {\em Proceedings of the 30th ACM Conference on Hypertext and
  Social Media}, pages 283--284, 2019.

\bibitem{Jain2017}
Atishay Jain and Faraz Wasim.
\newblock Answering squad.
\newblock {\em Department of Computer Science}, pages 94305--9020, 2017.

\bibitem{ravichandran2002learning}
Deepak Ravichandran and Eduard Hovy.
\newblock Learning surface text patterns for a question answering system.
\newblock In {\em Proceedings of the 40th Annual meeting of the association for
  Computational Linguistics}, pages 41--47, 2002.

\bibitem{antol2015vqa}
Stanislaw Antol, Aishwarya Agrawal, Jiasen Lu, Margaret Mitchell, Dhruv Batra,
  C~Lawrence~Zitnick, and Devi Parikh.
\newblock Vqa: Visual question answering.
\newblock In {\em Proceedings of the IEEE international conference on computer
  vision}, pages 2425--2433, 2015.

\bibitem{yang2016stacked}
Zichao Yang, Xiaodong He, Jianfeng Gao, Li~Deng, and Alex Smola.
\newblock Stacked attention networks for image question answering.
\newblock In {\em Proceedings of the IEEE conference on computer vision and
  pattern recognition}, pages 21--29, 2016.

\bibitem{kwiatkowski2019natural}
Tom Kwiatkowski, Jennimaria Palomaki, Olivia Redfield, Michael Collins, Ankur
  Parikh, Chris Alberti, Danielle Epstein, Illia Polosukhin, Jacob Devlin,
  Kenton Lee, et~al.
\newblock Natural questions: a benchmark for question answering research.
\newblock {\em Transactions of the Association for Computational Linguistics},
  7:453--466, 2019.

\bibitem{Rajpurkar2016}
Pranav Rajpurkar, Jian Zhang, Konstantin Lopyrev, and Percy Liang.
\newblock Squad: 100,000+ questions for machine comprehension of text.
\newblock {\em arXiv preprint arXiv:1606.05250}, 2016.

\bibitem{beli2003latent}
M~Beli~David, Y~Ng~Andrew, and I~Jordan~Michael.
\newblock Latent dirichlet allocation.
\newblock {\em Journal of machine Learning research}, 2003.

\bibitem{Liu2012}
Bing Liu.
\newblock Sentiment analysis and opinion mining.
\newblock {\em Synthesis lectures on human language technologies}, 5(1):1--167,
  2012.

\bibitem{lin2009joint}
Chenghua Lin and Yulan He.
\newblock Joint sentiment/topic model for sentiment analysis.
\newblock In {\em Proceedings of the 18th ACM conference on Information and
  knowledge management}, pages 375--384, 2009.

\bibitem{dos2014deep}
Cicero Dos~Santos and Maira Gatti.
\newblock Deep convolutional neural networks for sentiment analysis of short
  texts.
\newblock In {\em Proceedings of COLING 2014, the 25th International Conference
  on Computational Linguistics: Technical Papers}, pages 69--78, 2014.

\bibitem{bakshi2016opinion}
Rushlene~Kaur Bakshi, Navneet Kaur, Ravneet Kaur, and Gurpreet Kaur.
\newblock Opinion mining and sentiment analysis.
\newblock In {\em 2016 3rd International Conference on Computing for
  Sustainable Global Development (INDIACom)}, pages 452--455. IEEE, 2016.

\bibitem{Hutto2014}
Clayton~J Hutto and Eric Gilbert.
\newblock Vader: A parsimonious rule-based model for sentiment analysis of
  social media text.
\newblock In {\em Eighth international AAAI conference on weblogs and social
  media}, 2014.

\bibitem{ozbayoglu2020deep}
Ahmet~Murat Ozbayoglu, Mehmet~Ugur Gudelek, and Omer~Berat Sezer.
\newblock Deep learning for financial applications: A survey.
\newblock {\em Applied Soft Computing}, 93:106384, 2020.

\bibitem{necba2018using}
Hanae Necba, Maryem Rhanoui, and Bouchra~El Asri.
\newblock Using unsupervised machine learning for data quality. application to
  financial governmental data integration.
\newblock In {\em International Conference on Big Data, Cloud and
  Applications}, pages 197--209. Springer, 2018.

\bibitem{colladon2020forecasting}
A~Fronzetti Colladon, Stefano Grassi, Francesco Ravazzolo, and Francesco
  Violante.
\newblock Forecasting financial markets with semantic network analysis in the
  covid-19 crisis.
\newblock {\em arXiv preprint arXiv:2009.04975}, 2020.

\bibitem{Nassirtoussi2015}
Arman~Khadjeh Nassirtoussi, Saeed Aghabozorgi, Teh~Ying Wah, and David
  Chek~Ling Ngo.
\newblock Text mining of news-headlines for forex market prediction: A
  multi-layer dimension reduction algorithm with semantics and sentiment.
\newblock {\em Expert Systems with Applications}, 42(1):306--324, 2015.

\bibitem{Calomiris2019}
Charles~W Calomiris and Harry Mamaysky.
\newblock How news and its context drive risk and returns around the world.
\newblock {\em Journal of Financial Economics}, 133(2):299--336, 2019.

\bibitem{Tai2018}
Ichihan Tai.
\newblock {\em Multi-factor Sentiment Analysis for Gauging Investors Fear}.
\newblock PhD thesis, The George Washington University, 2018.

\bibitem{Wahyudin2016}
Irfan Wahyudin, Taufik Djatna, and Wisnu~Ananta Kusuma.
\newblock Cluster analysis for sme risk analysis documents based on pillar
  k-means.
\newblock {\em Telkomnika}, 14(2):674, 2016.

\end{thebibliography}

\end{document}